\documentclass{article}

\usepackage[utf8]{inputenc}
\usepackage[T1]{fontenc}
\usepackage{hyperref}
\usepackage{url}
\usepackage{booktabs}
\usepackage{amsfonts}
\usepackage{nicefrac}
\usepackage{microtype}
\usepackage{xcolor}
\usepackage{amsmath}
\usepackage{graphicx}
\usepackage{enumitem}
\usepackage{subcaption}
\usepackage{float}
\usepackage{algorithm}
\usepackage{algorithmic}
\usepackage{natbib}
\graphicspath{{./}}

\title{
Parent-Guided Semantic Reward Model (PGSRM): \\
Embedding-Based Reward Functions for Reinforcement Learning of Transformer Language Models
}

\author{
  Alexandr Plashchinsky \\  
  VECTOR Labs \\ 
  \texttt{aplashch@gmail.com}
}

\begin{document}

\maketitle

\begin{abstract}
We introduce the \textbf{Parent-Guided Semantic Reward Model (PGSRM)}, a lightweight reward framework for reinforcement learning (RL) of transformer language models. PGSRM replaces binary correctness signals, human preference data, and trained reward models with a simple signal: cosine similarity between a \emph{parent} model's reference output embedding and a \emph{child} model's generated output for the same input. This yields a dense, semantically meaningful reward with no human annotation or additional model training. We apply PGSRM on five language tasks and find that it produces smoother reward improvement and more stable PPO dynamics than a binary reward baseline, suggesting that embedding-based semantic rewards are a practical alternative to RLHF-style reward modeling for parent-guided alignment in smaller transformer models.
\end{abstract}

\section{Introduction}

Large language models (LLMs) are often trained and adapted with reinforcement learning (RL) to better align their behavior with human preferences and task-specific objectives. The most prominent example is Reinforcement Learning from Human Feedback (RLHF), where human annotators compare model outputs, a reward model is trained on these preferences, and the language model is optimized with Proximal Policy Optimization (PPO) \citep{christiano2017deep, ouyang2022training, stiennon2020learning}. Despite its success, RLHF requires substantial human feedback, a separate reward model, and remains vulnerable to reward hacking when that model mis-specifies the objective.

Reward design is challenging even in simpler RL settings. Binary or sparse correctness-based rewards provide little signal: most trajectories receive zero reward, and small quality improvements are invisible. This sparsity yields unstable training, slow learning, and premature convergence, as observed when using sparse sequence-level metrics (e.g., BLEU) as RL rewards for text generation \citep{ranzato2016sequence}. For language models with rich, high-dimensional outputs, collapsing performance into a single discrete reward discards most of the available information.

An alternative is to exploit \emph{semantic} information about model outputs. Modern embedding models map text into continuous vector spaces where cosine similarity or Euclidean distance reflects semantic relatedness. If a stronger ``parent'' model produces an ideal output for each input, the distance between the parent output and a weaker ``child'' model's output in embedding space provides a dense measure of alignment. This signal is richer than binary correctness, requires no human annotation, and avoids training a separate reward model.

We formalize this idea as the \textbf{Parent-Guided Semantic Reward Model (PGSRM)}, a reward framework for RL with transformer language models. For each prompt, a parent model (e.g., a GPT-4-class system) generates a reference response and a smaller child model (GPT-2 Small or GPT-2 Large) generates its own; both are embedded into a shared vector space (Numberbatch \citep{speer2017conceptnet} or \texttt{text-embedding-3-large} \citep{openai2023embeddings}), and their similarity is used as the reward for PPO optimization. This turns teacher--student alignment into an RL problem driven directly by semantic proximity, without explicit preference labels or a trained reward model.

We evaluate PGSRM on five tasks of increasing difficulty: (1) color mixing, (2) antonym generation, (3) word categorization, (4) exact-string copying, and (5) sentiment inversion. Across these settings, we compare PGSRM to a binary reward baseline that assigns a reward of $1$ to exactly correct outputs and $0$ otherwise. Our experiments show that PGSRM yields smoother reward improvement and more stable PPO metrics than the binary baseline, suggesting that embedding-based semantic rewards are a practical alternative to traditional correctness signals in language model RL.

\paragraph{Contributions.}
This paper makes the following contributions:
\begin{itemize}[noitemsep, topsep=2pt]
    \item We show that a simple embedding-based semantic reward yields smoother rewards, more stable PPO dynamics, and more consistent learning than a binary correctness baseline across five language tasks.
    \item We provide an empirical and qualitative analysis of this semantic reward framework, arguing that it offers a practical lightweight alternative to RLHF-style reward modeling for teacher--student alignment in language models.
\end{itemize}

\section{Related Work}

\paragraph{RLHF and AI feedback.}
Reinforcement Learning from Human Feedback (RLHF) is a standard approach for aligning large language models with human preferences: human annotators compare model outputs, a reward model is trained on these preferences, and the policy is optimized with PPO \citep{christiano2017deep, ouyang2022training, stiennon2020learning}. Despite its success, RLHF requires substantial human effort, introduces a potentially mis-specified reward model, and is vulnerable to reward hacking \citep{sharma2024critical}. Related approaches use feedback from stronger AI systems, e.g.\ constitutional RL \citep{bai2022constitutional}, but still rely on explicit scalar heuristics or trained reward models.

\paragraph{Teacher--student learning and distillation.}
Teacher--student frameworks and knowledge distillation transfer capabilities from large models to smaller ones by training a student to match a teacher's outputs or representations with supervised losses \citep{hinton2015distilling}. PGSRM is similar in spirit, but instead of supervised imitation it uses reinforcement learning with a scalar reward derived from semantic similarity between teacher and student outputs.

\paragraph{Embedding-based similarity and evaluation.}
Text embedding models are widely used for semantic search, clustering, reranking, and approximate evaluation, with cosine similarity serving as a proxy for semantic relatedness \citep{zhang2020bertscore}. Prior work mainly uses embeddings as diagnostic metrics or for retrieval, whereas PGSRM promotes embedding similarity to the \emph{reward function} in PPO. To our knowledge, parent--child embedding similarity has not previously been studied as a standalone reward for training transformer language models with RL.

\section{Method: Parent-Guided Semantic Reward Model (PGSRM)}
\label{sec:method}

PGSRM is a sequence-level RL framework in which a strong \emph{parent} model provides reference outputs, an embedding model maps text to a semantic vector space, and a weaker \emph{child} model is optimized with PPO using embedding-based similarity as the reward.

\subsection{Problem Setup}

Each input state $s$ is a text prompt, and the model must produce an output sequence $a$. We assume two language models: a fixed parent policy $\pi_p$ and a trainable child policy $\pi_\theta$. For each prompt $s$, the parent produces a reference output
\[
a_p = \pi_p(s),
\]
and the child produces its own output
\[
a_c \sim \pi_\theta(\cdot \mid s).
\]
We treat $(s, a_c)$ as a single-step episode with scalar reward $R(s, a_c, a_p)$ and update $\pi_\theta$ with PPO.

\subsection{Embedding-Based Reward}

To measure semantic similarity between outputs, we use a frozen text embedding model
$f: \mathcal{A} \rightarrow \mathbb{R}^d$, which maps a sequence to a fixed-dimensional vector.
For each prompt $s$, we embed both the parent and child outputs,
\[
e_p = f(a_p), \qquad e_c = f(a_c),
\]
and $\ell_2$-normalize them. Our primary similarity measure is cosine similarity,
\[
\mathrm{cos}(e_p, e_c) = e_p^\top e_c \in [-1, 1],
\]
which we truncate and optionally sharpen:
\[
R_{\text{PGSRM}}(s, a_c, a_p) = \bigl(\max(0, \mathrm{cos}(e_p, e_c))\bigr)^\alpha,
\]
where $\alpha \ge 1$ controls how strongly we emphasize high-similarity outputs. This yields a dense, non-negative scalar reward for every child output, with higher reward for outputs that are closer to the parent in embedding space.

\subsection{PPO Training}

We optimize the child policy $\pi_\theta$ with Proximal Policy Optimization (PPO) \citep{schulman2017proximal}, treating $R_{\text{PGSRM}}$ as the episode return. For each batch of prompts, we sample child outputs, compute their PGSRM rewards, and estimate advantages with a learned value function $V_\phi(s)$. We then update $\pi_\theta$ using the standard clipped PPO objective with a light KL penalty between successive policies. In most experiments we use minimal additional regularization so that the policy can respond directly to small but consistent differences in semantic reward. Full optimization details and hyperparameters are provided in the appendix.

\section{Experimental Setup}
\label{sec:experimental_setup}

We evaluate PGSRM using GPT-2 child policies on five tasks of increasing difficulty, comparing against a binary correctness reward while keeping the RL pipeline and hyperparameters fixed.

\subsection{Models}

We use GPT-2 Small (124M parameters) for the three simpler tasks (color mixing, antonym generation, word categorization) and GPT-2 Large (774M parameters) for the two more complex tasks (exact-string copying and sentiment inversion) \citep{radford2019language}. In all cases, the child starts from the publicly available pre-trained checkpoint and is fine-tuned with PPO. For all tasks, the parent policy $\pi_p$ is a GPT-4-class model (\texttt{gpt-4o-mini}) that we treat as an oracle: for each prompt $s$ in a task’s dataset, we query the parent once offline to obtain a reference output $a_p = \pi_p(s)$, and the parent is frozen during training. The embedding function $f$ is instantiated as \textbf{Numberbatch} for color mixing, antonym generation, and word categorization, and as \textbf{\texttt{text-embedding-3-large}} for exact copying and sentiment inversion. In all experiments, $f$ is frozen and embeddings are $\ell_2$-normalized before computing cosine similarity.

\subsection{Tasks}

We evaluate PGSRM on five parent–child alignment tasks:

\begin{itemize}[noitemsep, topsep=2pt]
    \item \textbf{Color mixing:} $s$ is a textual prompt of two basic colors. $a_p$ is the resulting color name. The child must output the correct mixture color.
    \item \textbf{Antonym generation:} $s$ is a short template containing an adjective. $a_p$ is an appropriate antonym. The child is rewarded for producing the correct antonym.
    \item \textbf{Word categorization:} $s$ is a noun or short phrase. $a_p$ is a coarse semantic category. The task probes mapping concrete entities to high-level classes.
    \item \textbf{Exact-string copying:} $s$ is a full text sequence, and $a_p$ is an exact copy of $s$. The child’s goal is to reproduce the full sequence.
    \item \textbf{Sentiment inversion:} $s$ is a first-person sentence expressing positive sentiment. $a_p$ is a semantically related first-person sentence with negative sentiment. The child's goal is to produce this inverted sentiment sentence.
\end{itemize}

\subsection{Baseline}

As a baseline, we use a \textbf{binary correctness reward} for each task. An output is treated as correct if it satisfies a task-specific checker (e.g., string equality after simple normalization for color mixing, antonyms, categorization, and copying; a rule-based template check for sentiment inversion). Correct outputs receive reward $1$ and all others receive reward $0$. The policy architecture, optimizer, and PPO configuration are identical between PGSRM and the binary baseline; only the reward function differs.

\subsection{Training Configuration}

\paragraph{Episodes and batch sizes.}
For each task and each reward type (PGSRM vs. binary), we train for \textbf{100{,}000} single-step PPO episodes. Each episode consists of sampling one prompt $s$, generating one child output $a_c$, computing its scalar reward, and including that sample in the PPO update. We use:
\begin{itemize}[noitemsep, topsep=2pt]
    \item \textbf{Batch size 50} for color mixing, antonyms, and categorization;
    \item \textbf{Batch size 10} for exact copying and sentiment inversion.
\end{itemize}
In all cases, the total number of episodes per run is fixed at 100{,}000.

\paragraph{Optimization hyperparameters.}
We use separate optimizers for the policy and value function with shared settings across all tasks and both reward types:
\begin{itemize}[noitemsep, topsep=2pt]
    \item Actor learning rate: $1 \times 10^{-5}$;
    \item Critic learning rate: $1 \times 10^{-4}$;
    \item Entropy coefficient: $0.01$;
    \item Value loss coefficient: $0.5$;
    \item Initial KL penalty coefficient: $5 \times 10^{-5}$;
    \item Max gradient norm: $1.0$.
\end{itemize}

\paragraph{PPO regularization.}
We train with PPO \citep{schulman2017proximal} but weaken some stabilizers so the policy can react more directly to the dense PGSRM reward. Specifically, we omit ratio clipping from the policy loss and rely on a light KL penalty between old and new policies. For the three simpler tasks, we use an adaptive KL coefficient targeting a moderate KL divergence; for the two more complex tasks, we fix the KL coefficient at $5 \times 10^{-5}$. Full hyperparameter details are provided in the appendix.

\section{Results}
\label{sec:results}

\begin{figure}[H]
    \centering
    \begin{subfigure}[t]{0.32\textwidth}
        \includegraphics[width=\linewidth]{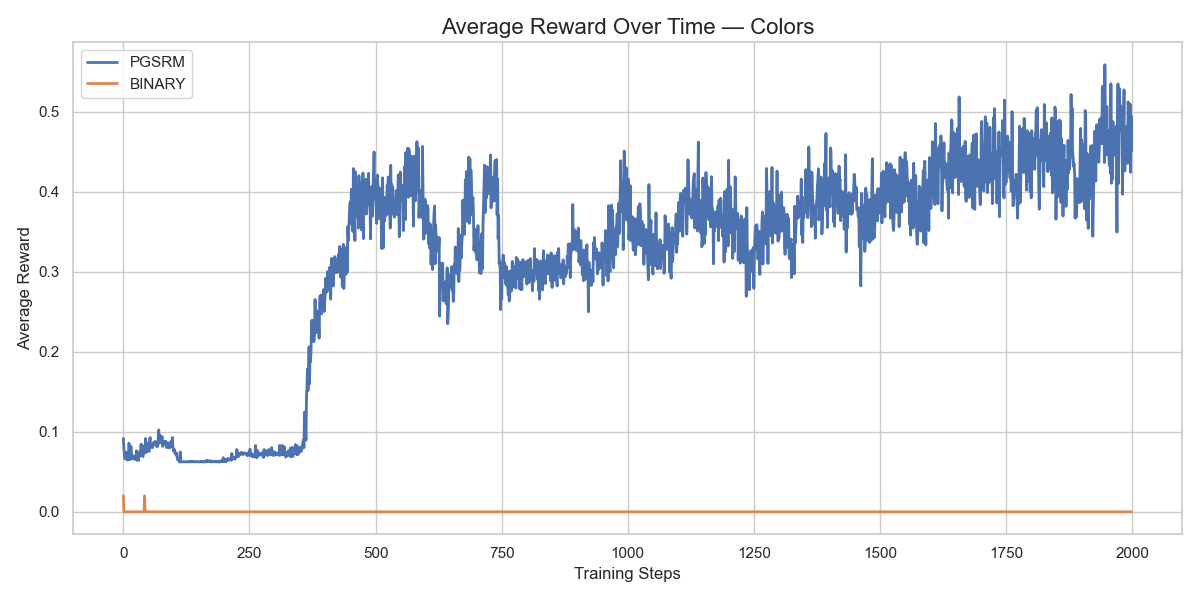}
        \caption{Average reward}
    \end{subfigure}
    \hfill
    \begin{subfigure}[t]{0.32\textwidth}
        \includegraphics[width=\linewidth]{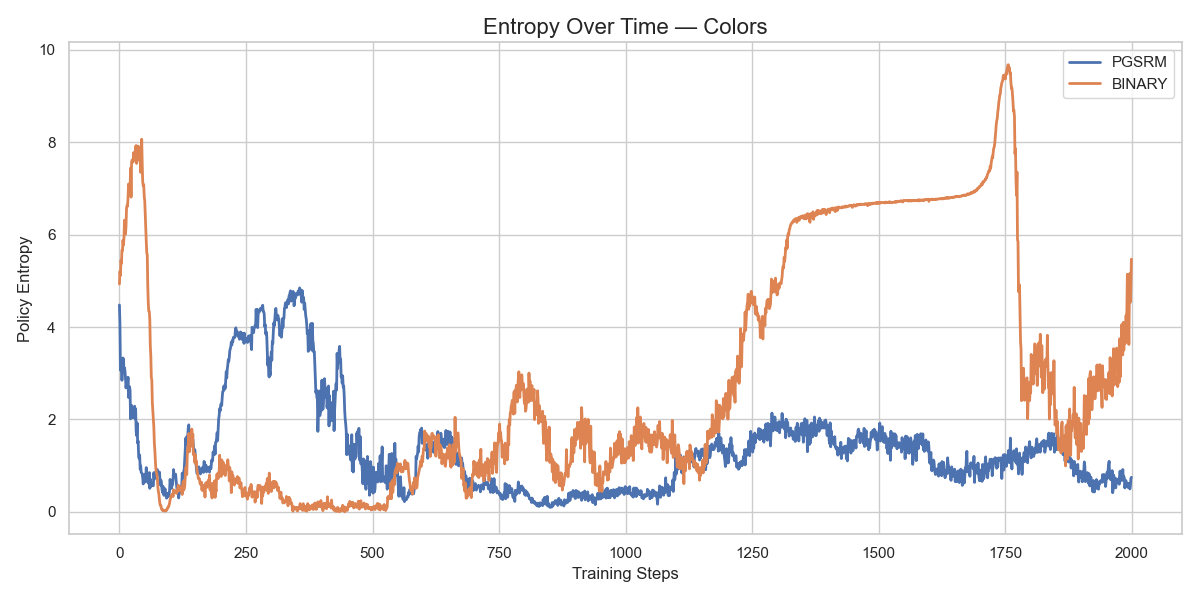}
        \caption{Entropy}
    \end{subfigure}
    \hfill
    \begin{subfigure}[t]{0.32\textwidth}
        \includegraphics[width=\linewidth]{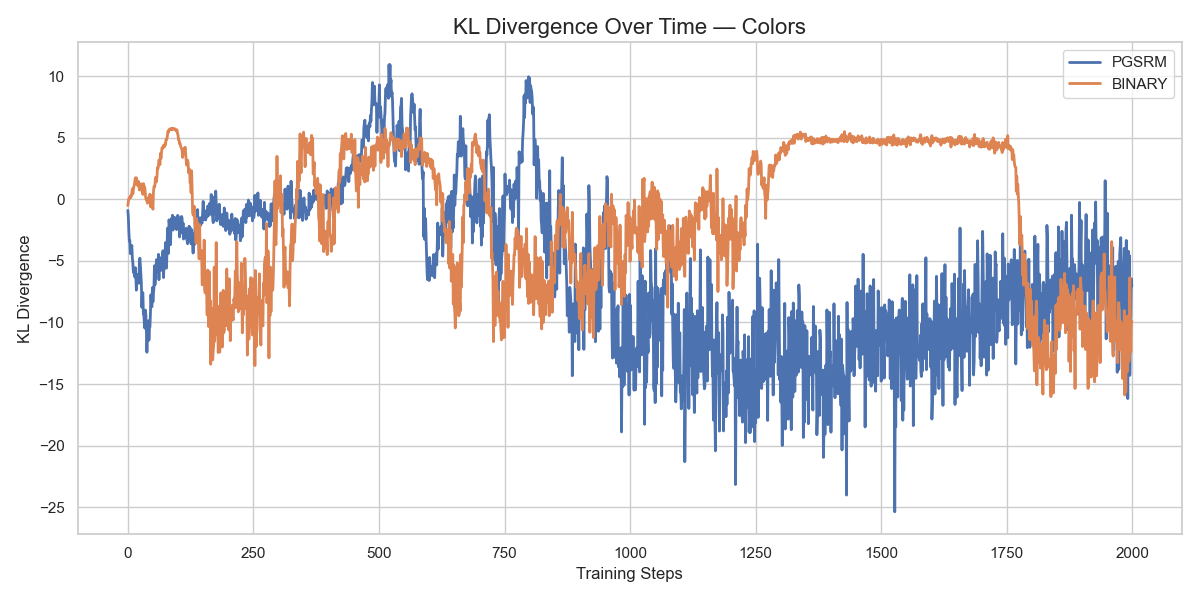}
        \caption{KL divergence}
    \end{subfigure}

    \caption{Training dynamics for the color mixing task.}
    \label{fig:results-colors}
\end{figure}

\begin{figure}[H]
    \centering
    \begin{subfigure}[t]{0.32\textwidth}
        \includegraphics[width=\linewidth]{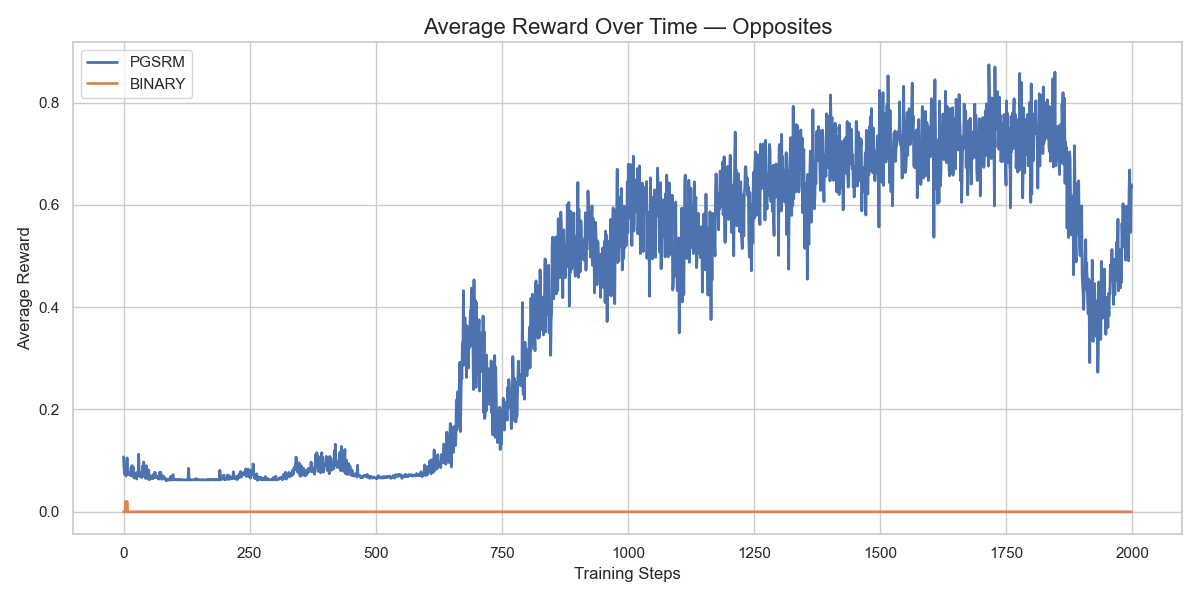}
        \caption{Average reward}
    \end{subfigure}
    \hfill
    \begin{subfigure}[t]{0.32\textwidth}
        \includegraphics[width=\linewidth]{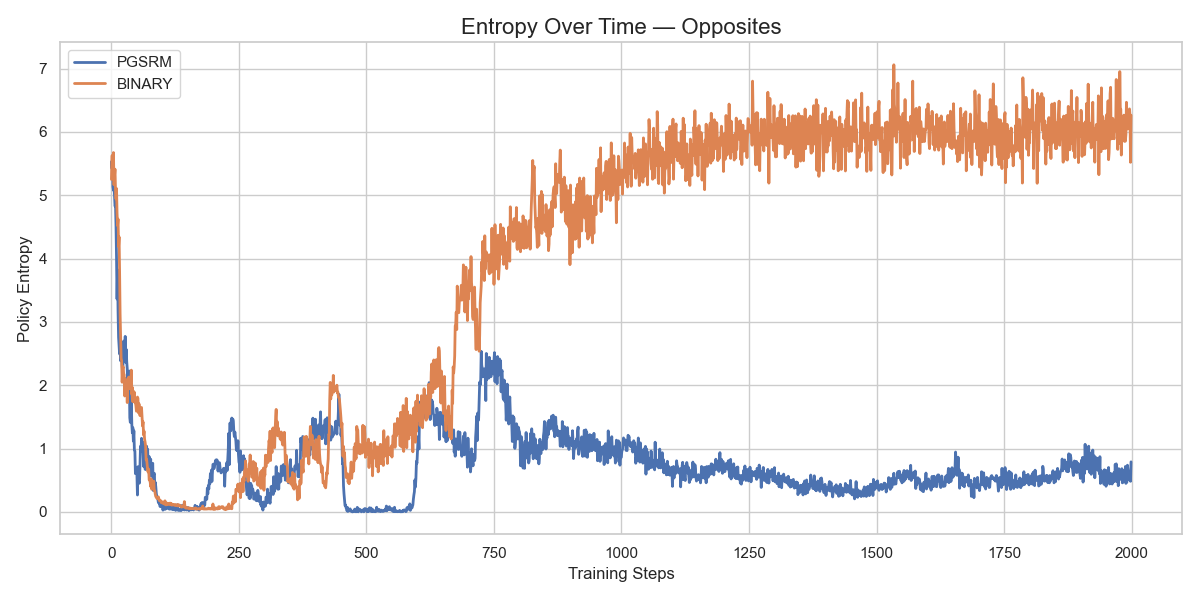}
        \caption{Entropy}
    \end{subfigure}
    \hfill
    \begin{subfigure}[t]{0.32\textwidth}
        \includegraphics[width=\linewidth]{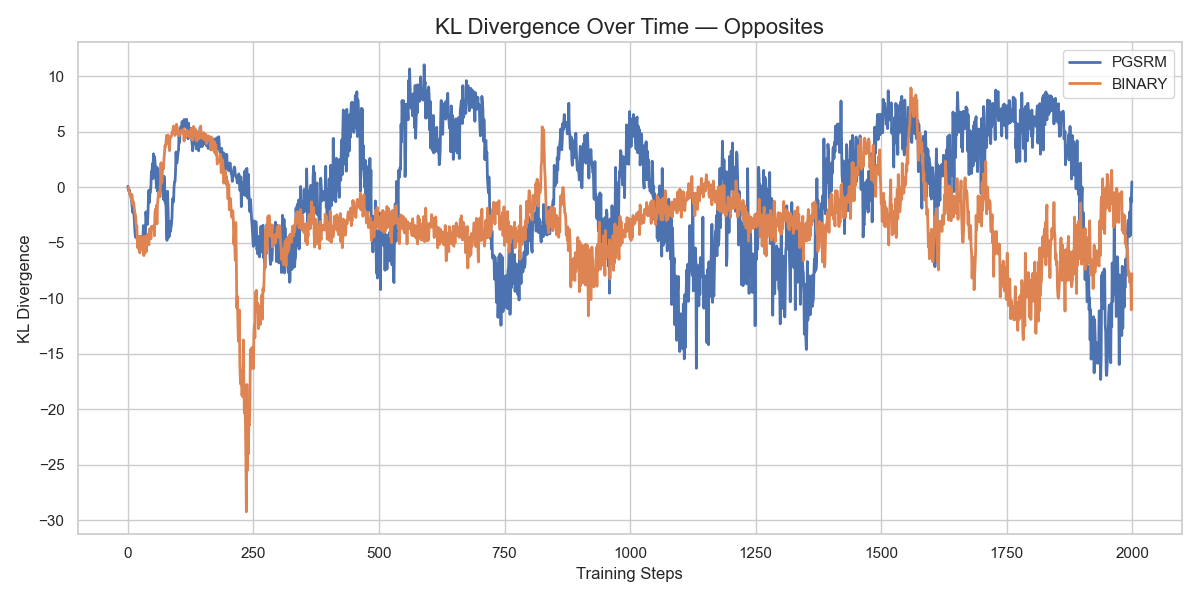}
        \caption{KL divergence}
    \end{subfigure}

    \caption{Training dynamics for the antonym generation task.}
    \label{fig:results-opposites}
\end{figure}

\begin{figure}[H]
    \centering
    \begin{subfigure}[t]{0.32\textwidth}
        \includegraphics[width=\linewidth]{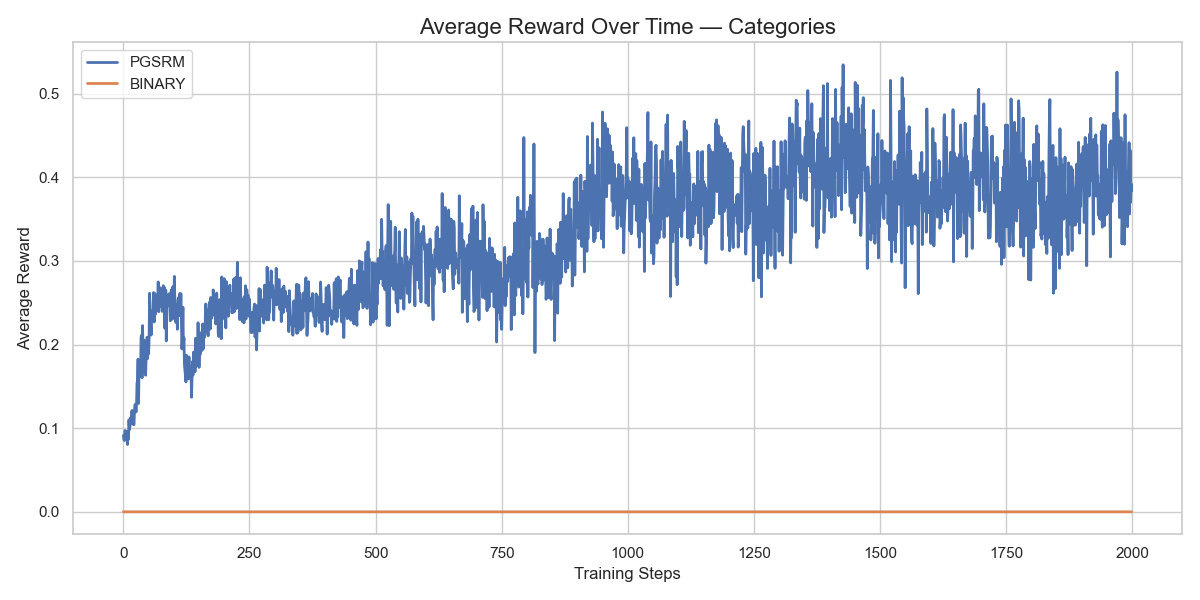}
        \caption{Average reward}
    \end{subfigure}
    \hfill
    \begin{subfigure}[t]{0.32\textwidth}
        \includegraphics[width=\linewidth]{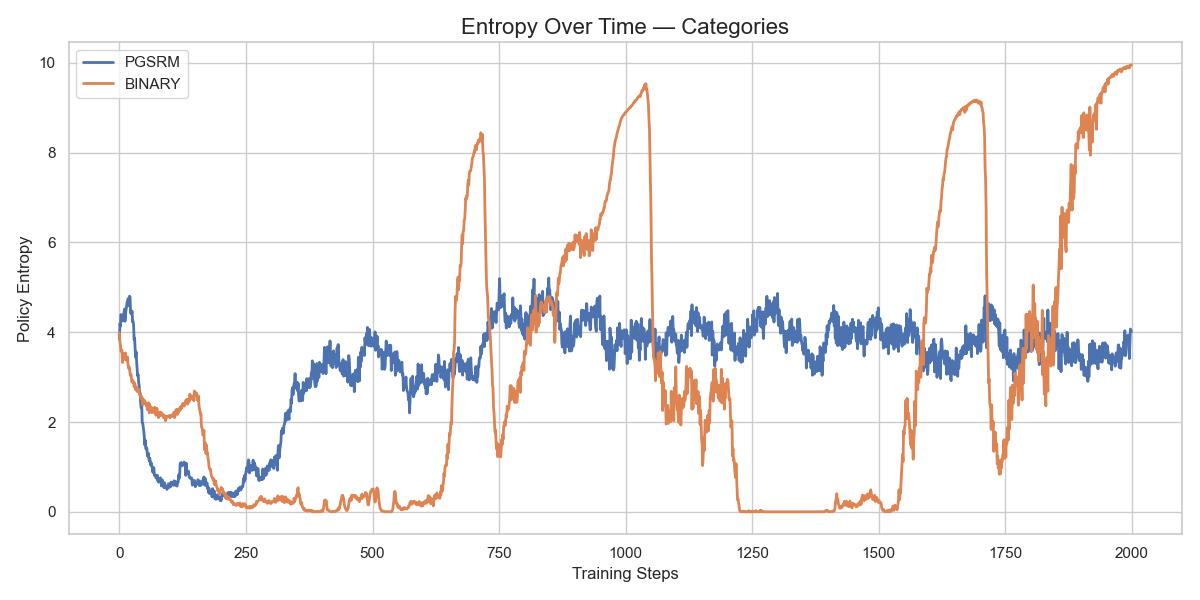}
        \caption{Entropy}
    \end{subfigure}
    \hfill
    \begin{subfigure}[t]{0.32\textwidth}
        \includegraphics[width=\linewidth]{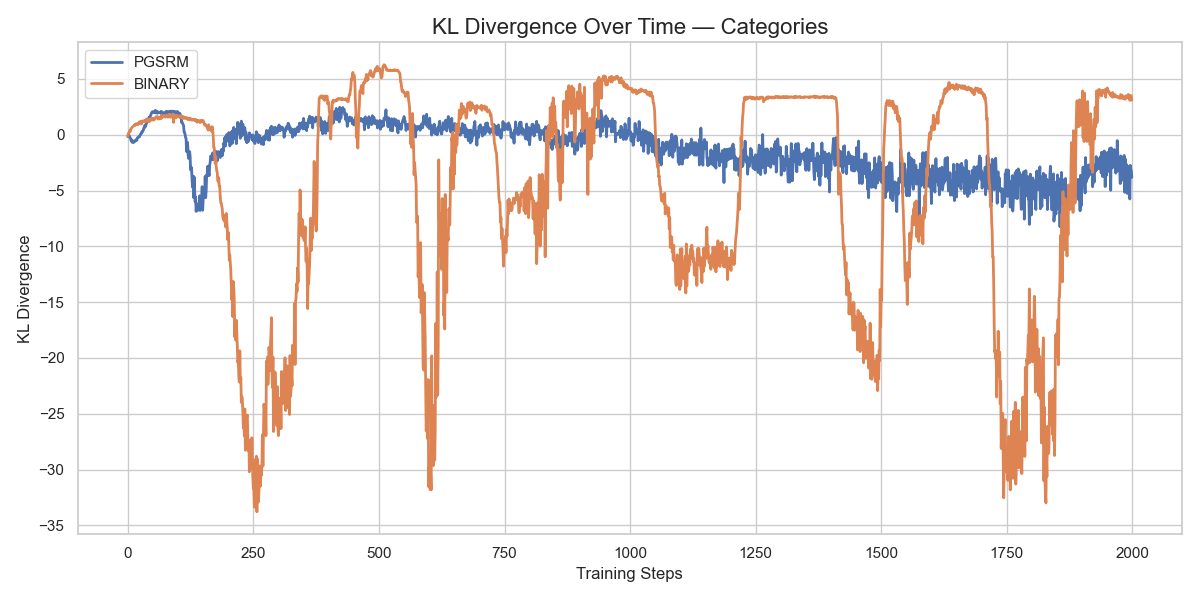}
        \caption{KL divergence}
    \end{subfigure}

    \caption{Training dynamics for the word categorization task.}
    \label{fig:results-categories}
\end{figure}

\begin{figure}[H]
    \centering
    \begin{subfigure}[t]{0.32\textwidth}
        \includegraphics[width=\linewidth]{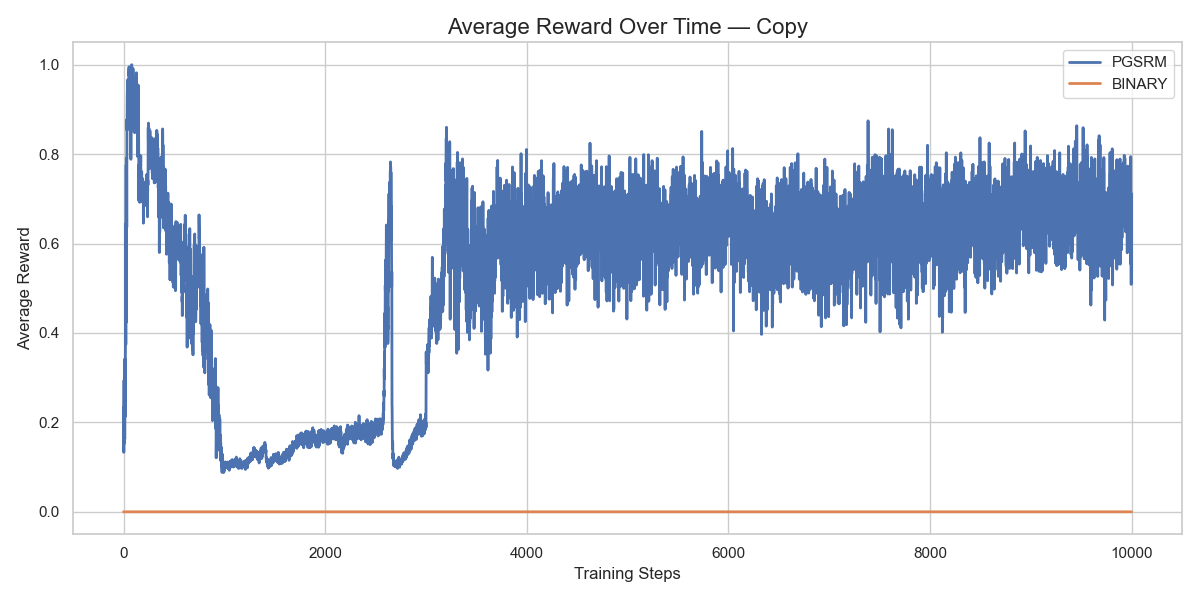}
        \caption{Average reward}
    \end{subfigure}
    \hfill
    \begin{subfigure}[t]{0.32\textwidth}
        \includegraphics[width=\linewidth]{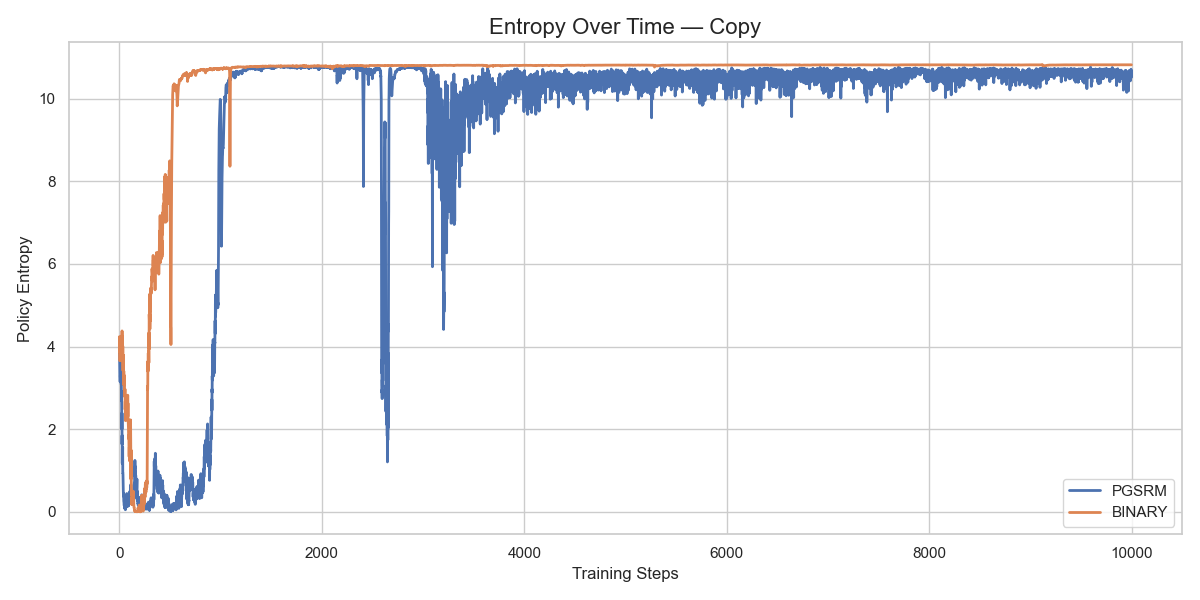}
        \caption{Entropy}
    \end{subfigure}
    \hfill
    \begin{subfigure}[t]{0.32\textwidth}
        \includegraphics[width=\linewidth]{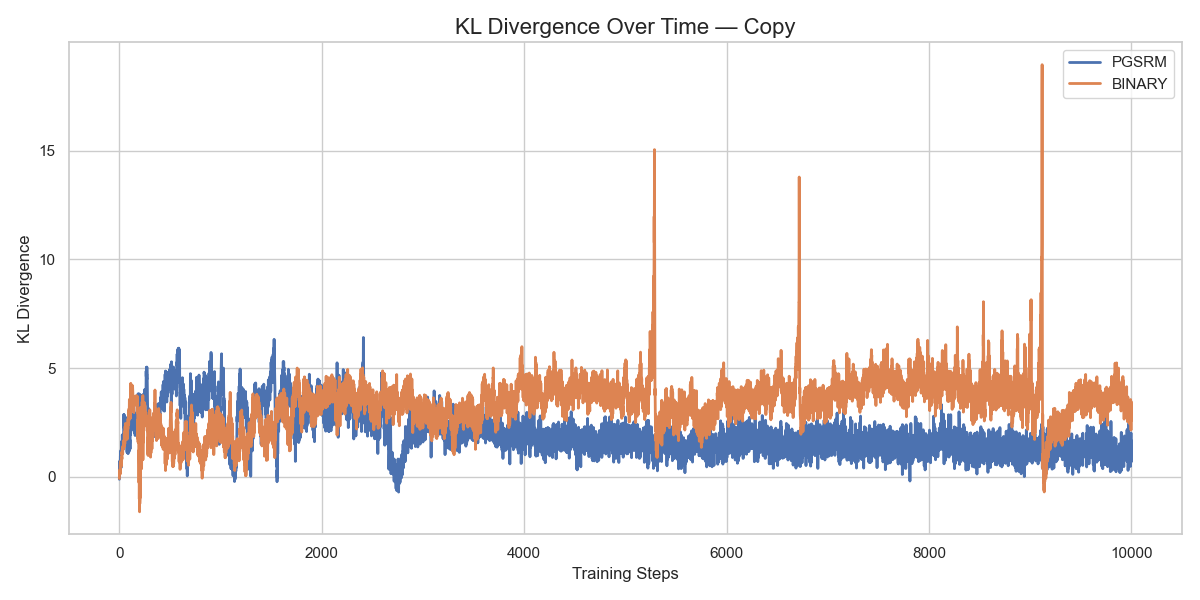}
        \caption{KL divergence}
    \end{subfigure}

    \caption{Training dynamics for the exact-string copying task.}
    \label{fig:results-copy}
\end{figure}

\begin{figure}[H]
    \centering
    \begin{subfigure}[t]{0.32\textwidth}
        \includegraphics[width=\linewidth]{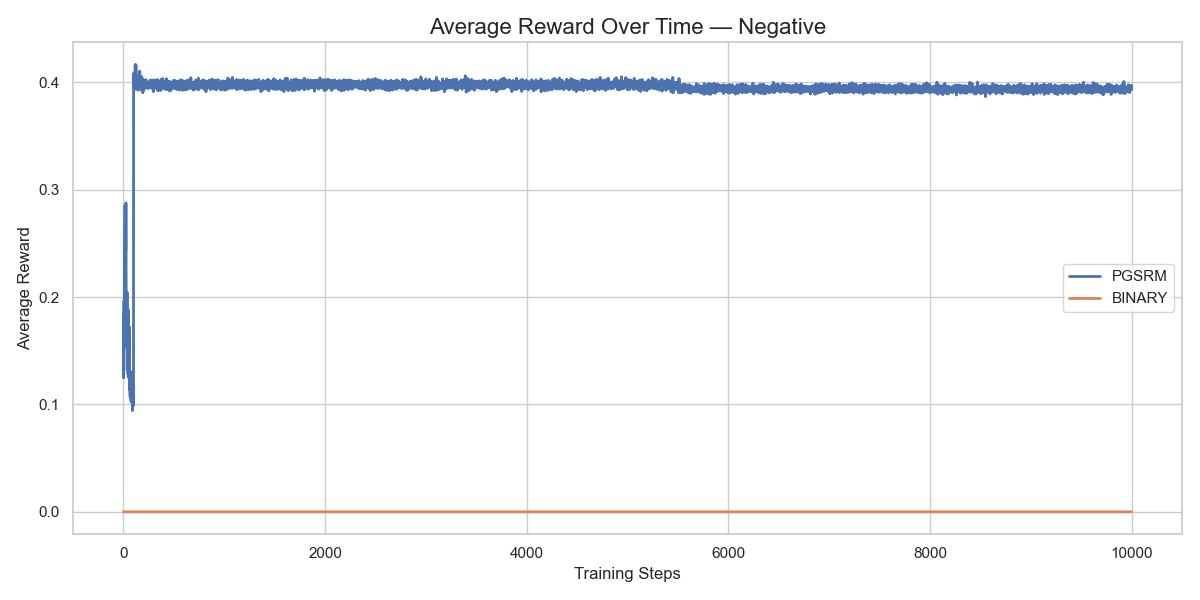}
        \caption{Average reward}
    \end{subfigure}
    \hfill
    \begin{subfigure}[t]{0.32\textwidth}
        \includegraphics[width=\linewidth]{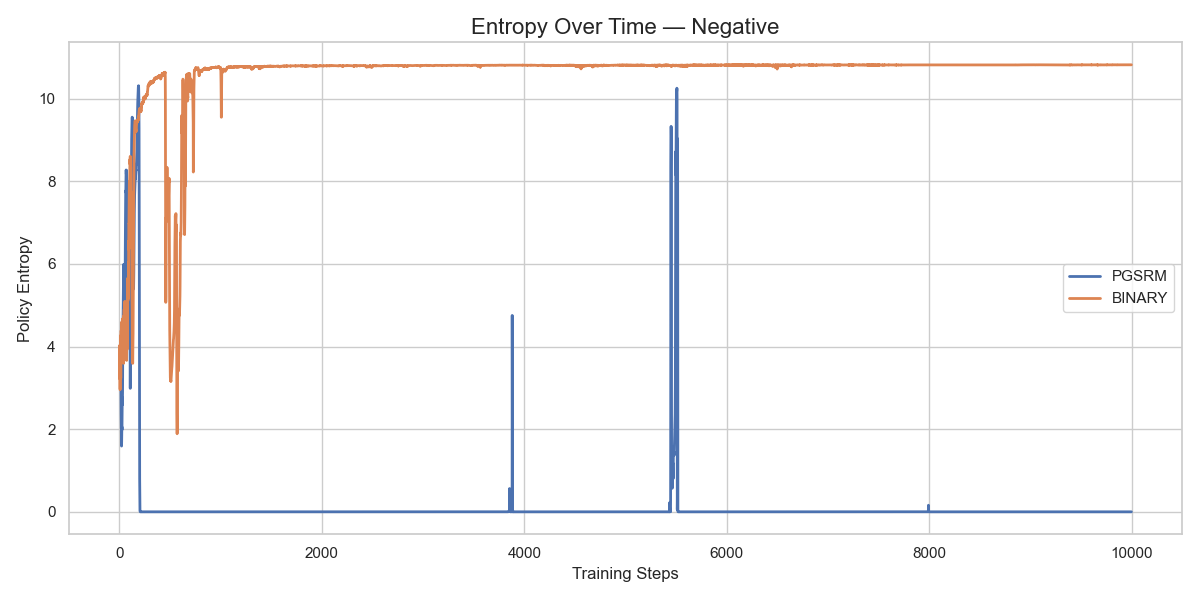}
        \caption{Entropy}
    \end{subfigure}
    \hfill
    \begin{subfigure}[t]{0.32\textwidth}
        \includegraphics[width=\linewidth]{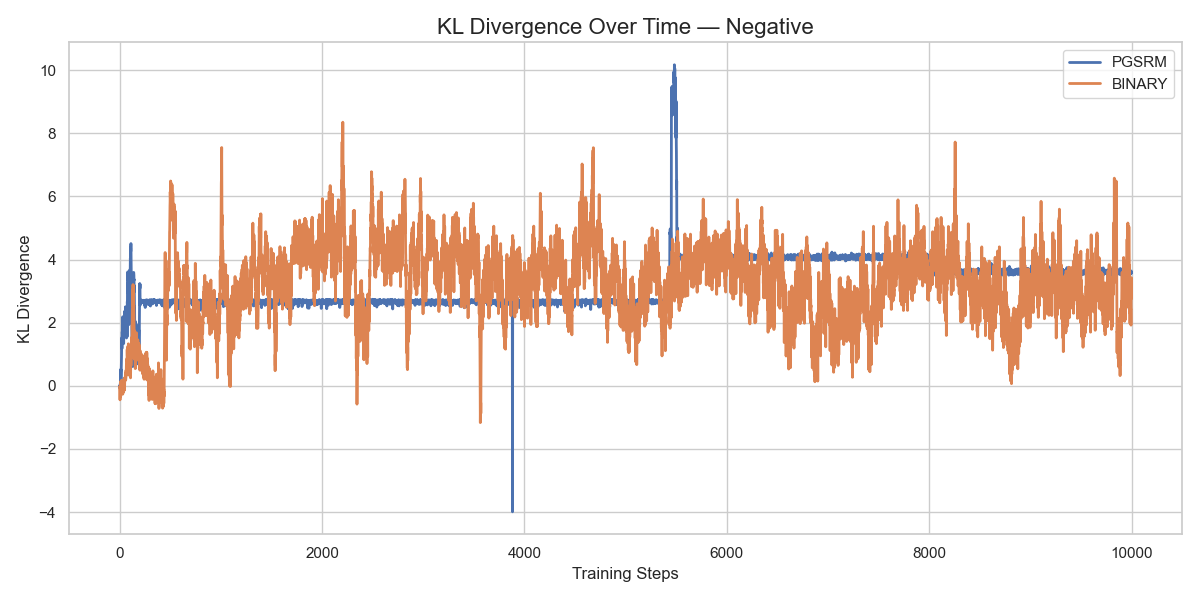}
        \caption{KL divergence}
    \end{subfigure}

    \caption{Training dynamics for the sentiment inversion task.}
    \label{fig:results-negative}
\end{figure}

We now present empirical results comparing PGSRM against a binary correctness reward across the five tasks described in Section~\ref{sec:experimental_setup}. For each task, we visualize three key training metrics: (i) average reward, (ii) policy entropy, and (iii) KL divergence between successive policies. Figures~\ref{fig:results-colors}--\ref{fig:results-negative} show these trajectories for PGSRM and the binary baseline side by side. In what follows, we analyze each task in turn.

\subsection{Color Mixing}

PGSRM exhibits a clear learning signal on color mixing: after an initial low-reward phase, the average reward undergoes a sharp jump and then continues to drift upward, indicating that the GPT-2 Small child discovers a useful color-mapping strategy and refines it over time. The binary baseline remains essentially flat at (or near) zero reward, apart from a few brief spikes, since exact matches are rare and semantically close guesses receive no credit.

The PPO diagnostics mirror this pattern. Under PGSRM, entropy quickly drops from a high initial value, then recovers to a moderate, stable band, and KL divergence stays in a bounded range, suggesting controlled but effective policy updates. Under binary reward, entropy oscillates between near-collapse and high diffusion, and KL alternates between plateaus and sharp jumps, consistent with unstable updates driven by sparse, poorly informative feedback.

\subsection{Antonym Generation}

For antonym generation, PGSRM again yields a clear learning signal. The average reward starts at a low but nontrivial level, then undergoes a sharp increase mid-training and approaches a high plateau (around 0.7--0.8), indicating that GPT-2 Small learns increasingly reliable opposite mappings. The binary baseline remains essentially flat at (or near) zero reward, apart from a few small spikes, because exact matches between child and parent outputs are too rare to provide sustained positive feedback within 100{,}000 episodes.

Entropy and KL divergence show how the two rewards shape exploration. Under PGSRM, entropy quickly collapses from a high initial value, briefly re-expands, and then settles into a low but nonzero band, consistent with a policy that first commits to a pattern of antonyms, then refines it while retaining some stochasticity. The binary run, by contrast, spends most of training at high entropy with small, hesitant KL changes, reflecting diffuse, unguided exploration that does not translate into reward improvement.

\subsection{Word Categorization}

For categorization, PGSRM yields a steady, nontrivial learning signal: average reward rises from about 0.1 at the start of training into the 0.35–0.5 range by the end, indicating that the child policy both discovers useful category mappings and continues to refine them over time. The binary baseline remains essentially flat at zero reward, as exact category matches are too rare under a strict $\{0,1\}$ criterion to generate informative gradients within 100{,}000 episodes.

Entropy and KL divergence again reflect this contrast. Under PGSRM, entropy quickly drops from its initial value and then stabilizes in a moderate band, suggesting a balanced policy that is confident but still explores. Under binary reward, entropy repeatedly crashes toward zero and then spikes toward its maximum, accompanied by large KL excursions, indicating unstable oscillations between collapsed and nearly random behavior with no corresponding improvement in reward.

\subsection{Exact-String Copying}

In the copying task, PGSRM produces a clear phase transition: after an initial transient with a brief reward spike and a low–medium plateau, the average reward jumps into a high band (around 0.6--0.8) and remains there, indicating that GPT-2 Large learns a robust copying strategy across the prompt set. The binary baseline stays essentially flat at zero reward throughout, since exact string matches are extremely rare and near-miss outputs receive no credit under a $\{0,1\}$ signal.

Entropy and KL show how PGSRM still drives learning in a high-entropy regime. By the end of training, both PGSRM and binary runs exhibit high entropy, but only PGSRM converts this exploration into consistently high reward. Under PGSRM, KL divergence remains in a moderate, stable range despite the absence of adaptive KL control, while the binary baseline shows occasional large KL spikes that do not translate into sustained reward gains.

\subsection{Sentiment Inversion}

In the sentiment inversion task, PGSRM does not produce a slow, gradual climb but instead a rapid jump followed by a stable plateau: after a brief transient, the average reward quickly rises to around 0.4 and then remains roughly constant, suggesting that GPT-2 Large quickly discovers a family of negative rewrites that score well under the semantic reward and exploits them for the rest of training. The binary baseline stays essentially flat at zero reward, since exact template-level matches to the parent’s inverted sentence are extremely rare, highlighting the importance of dense semantic feedback in this high-precision sequence transformation setting.

The PPO diagnostics are consistent with this picture. Under PGSRM, entropy briefly increases during early exploration but then collapses toward zero and stays low, indicating a highly deterministic rewrite strategy, while KL divergence remains in a moderate band with occasional spikes corresponding to small shifts in the dominant pattern. Under binary reward, entropy quickly rises to very high values and remains there, and KL is noisier and more erratic, reflecting unguided exploration that moves the policy in parameter space without yielding meaningful reward gains.

\section{Discussion and Limitations}

Across five qualitatively different tasks and two model sizes, PGSRM consistently produces non-trivial reward improvement and stable PPO dynamics, while a binary correctness baseline mostly stagnates near zero reward or exhibits unstable behavior. The main driver is \emph{reward density}: every output receives a graded signal based on embedding similarity to the parent, so almost every update carries information about how much better or worse the behavior is. Partial matches in color mixing, antonyms, categorization, copying, and sentiment inversion all yield intermediate scores, producing smoother advantages and more directional policy gradients, with phase transitions in reward and bounded KL divergence even under weak regularization.

Embedding-based reward also acts as a simple form of \emph{semantic shaping}. Instead of rewarding a single surface form, PGSRM encourages the child to move into the semantic neighborhood of the parent’s output. Near-miss responses receive partial credit rather than being collapsed to zero, effectively defining a smooth reward manifold around teacher behavior and allowing PPO to take relatively large steps without diverging.

PGSRM also has important limitations. It is fundamentally a teacher-imitation objective: in expectation the child cannot systematically outperform the parent in the embedding space and will reproduce any systematic errors, biases, or safety issues present in the parent. Embedding similarity is only a proxy for true task reward, especially for long-form text, and can be exploited via ``reward hacking'' where generic or templated responses score well without fully satisfying downstream constraints. Our experiments are restricted to short, single-step tasks with synthetic or simplified structure and GPT-2–scale models, without long-horizon credit assignment, tool use, or human evaluation. Finally, because PGSRM relies on a single off-the-shelf embedding model, it inherits that model’s inductive biases and blind spots, so any real-world alignment or safety use would require additional auditing and evaluation beyond the scope of this work.

\section{Conclusion}

We introduced PGSRM, an embedding-based reward scheme that turns a frozen teacher model into a semantic reward function for reinforcement learning. Instead of sparse binary correctness, PGSRM scores each child output by its embedding similarity to the teacher’s response, providing a dense, task-agnostic signal that can be plugged into standard PPO without training a separate reward model.

Across five diverse tasks and two GPT-2 sizes, PGSRM yields higher rewards and more stable PPO diagnostics than a binary correctness baseline, suggesting that shaping the reward landscape in embedding space can substantially improve optimization stability and sample efficiency. PGSRM offers a lightweight alternative to full RLHF-style reward modeling for teacher–student RL and distillation of specialized models, but it also inherits the biases and limitations of its teacher and embedding space. It is therefore best viewed as a useful building block within a broader alignment toolkit rather than a standalone solution.

\clearpage
\bibliographystyle{plainnat}
\bibliography{references}

\clearpage
\appendix

\section{Additional Experimental Details}
\label{app:details}

\subsection{Code and Environment}
\label{app:code}

All experiments were implemented in Python using a standard transformer RL stack.\footnote{We release the full training code, configuration files, and prompt datasets at \url{https://github.com/alexplash/project_adam}.}%
The PGSRM reward, binary baseline, and PPO training loop share a single codebase; the only difference between conditions is the reward computation (semantic similarity vs.\ binary correctness).

\paragraph{Hardware.}
All runs were executed on rented GPUs (e.g., NVIDIA A100/RTX-class GPUs) with at least 40~GB of VRAM.\footnote{Exact hardware is not required to reproduce the qualitative trends; any modern GPU with sufficient memory to fine-tune GPT-2 Large should suffice.}
We did not retain model checkpoints after training; the released code allows independent re-training from the same initialization and hyperparameters.

\subsection{Hyperparameters by Task}
\label{app:hyperparams}

Tables~\ref{tab:hp-simple} and \ref{tab:hp-complex} summarize the key hyperparameters for the simpler (GPT-2 Small) and more complex (GPT-2 Large) tasks, respectively. Unless otherwise noted, PGSRM and the binary baseline share identical settings; only the reward definition differs.

\begin{table}[H]
    \centering
    \caption{Hyperparameters for color mixing, antonym generation, and word categorization.}
    \label{tab:hp-simple}
    \begin{tabular}{lcc}
        \toprule
        Hyperparameter & Value & Notes \\
        \midrule
        Model & GPT-2 Small & 124M parameters \\
        Episodes & 100{,}000 & Single-step bandit episodes \\
        Batch size & 50 & Episodes per PPO update \\
        Actor learning rate & $1 \times 10^{-5}$ & \\
        Critic learning rate & $1 \times 10^{-4}$ & \\
        Entropy coefficient & 0.01 & \\
        Value loss coefficient & 0.5 & \\
        Initial KL coefficient & $5 \times 10^{-5}$ & Adaptive \\
        Target KL & 0.8 & Adaptive schedule (see main text) \\
        Gradient clipping & 1.0 & Max grad norm \\
        Embedding model & Numberbatch & $\ell_2$-normalized \\
        Reward type (PGSRM) & Cosine similarity$^\alpha$ & \\
        Cosine exponent $\alpha$ & 4 & Fixed for all runs \\
        Reward type (baseline) & Binary $\{0, 1\}$ & String equality \\
        \bottomrule
    \end{tabular}
\end{table}

\begin{table}[H]
    \centering
    \caption{Hyperparameters for exact-string copying and sentiment inversion.}
    \label{tab:hp-complex}
    \begin{tabular}{lcc}
        \toprule
        Hyperparameter & Value & Notes \\
        \midrule
        Model & GPT-2 Large & 774M parameters \\
        Episodes & 100{,}000 & Single-step bandit episodes \\
        Batch size & 10 & Episodes per PPO update \\
        Actor learning rate & $1 \times 10^{-5}$ & \\
        Critic learning rate & $1 \times 10^{-4}$ & \\
        Entropy coefficient & 0.01 & \\
        Value loss coefficient & 0.5 & \\
        KL coefficient & $5 \times 10^{-5}$ & Fixed (no adaptation) \\
        Gradient clipping & 1.0 & Max grad norm \\
        Embedding model & \texttt{text-embedding-3-large} & $\ell_2$-normalized \\
        Reward type (PGSRM) & Cosine similarity$^\alpha$ & \\
        Cosine exponent $\alpha$ & 4 & Fixed for all runs \\
        Reward type (baseline) & Binary $\{0, 1\}$ & Exact / template checks \\
        \bottomrule
    \end{tabular}
\end{table}

\subsection{Prompt Templates and Dataset Construction}
\label{app:prompts}

For reproducibility, we document the instruction prompts used for the parent (GPT-4-class) model and the corresponding prompts used for the child (GPT-2) policy on each task. In all cases, the parent receives a fixed natural-language instruction plus the task input, and the child is queried with a shorter, task-specific template. All randomness comes from sampling the child model; parent outputs are generated deterministically with temperature $= 0$.

\paragraph{Color mixing.}

\textbf{Parent instruction prompt (PGSRM target).}
\begin{verbatim}
You are an AI assistant tasked with combining two colors and
outputting the resulting color name.
For example:
- "red + blue = " → "purple"
- "red + yellow = " → "orange"
- "blue + yellow = " → "green"
- "white + black = " → "gray"
If the colors are similar (e.g. "red + pink"), output the most
dominant or blended color (e.g. "light red" or "pink").
Always output a single lowercase color name or a simple descriptive
blend like "light blue" or "dark green".
\end{verbatim}

For a given pair of colors $(c_1, c_2)$, the content string passed to the parent is:
\begin{verbatim}
"c1 + c2 ="
\end{verbatim}

\textbf{Child prompt.}
For the same pair $(c_1, c_2)$, the child is queried with:
\begin{verbatim}
"Mix the two colors and output only one lowercase color name:
 c1 and c2 is => "
\end{verbatim}

\paragraph{Antonym generation.}

\textbf{Parent instruction prompt (PGSRM target).}
\begin{verbatim}
You are an AI assistant that outputs the opposite (antonym) of
a given word.

You will receive an INPUT containing a single English word,
and you must respond with its opposite meaning.

Examples:
- "hot" → "cold"
- "up" → "down"
- "happy" → "sad"
- "light" → "dark"
- "old" → "young"
- "empty" → "full"

Rules:
- Output only a single lowercase word.
- No punctuation, no explanations, no extra words.
- Output must appear as {"OUTPUT": "<word>"} as required.
\end{verbatim}

\textbf{Child prompt.}
For an adjective \texttt{word}, the child prompt is:
\begin{verbatim}
"The opposite of 'word' is "
\end{verbatim}

\paragraph{Word categorization.}

\textbf{Parent instruction prompt (PGSRM target).}
\begin{verbatim}
You are an AI assistant that identifies the correct category of
a given noun.

Your job:
- Read the input word.
- Output the category it belongs to.
- Use a single, simple lowercase category word.

Examples:
- "apple" → "fruit"
- "tiger" → "animal"
- "rose" → "flower"
- "guitar" → "instrument"
- "carrot" → "vegetable"
- "eagle" → "bird"
- "shirt" → "clothing"

Rules:
- Output only one lowercase category word.
- Do NOT return the original word.
- No punctuation, no explanations.
- Output must appear as {"OUTPUT": "<category>"} as required.
\end{verbatim}

\textbf{Child prompt.}
For a noun \texttt{word}, the child is queried with:
\begin{verbatim}
"What category does this word belong to: 'word'?"
\end{verbatim}

\paragraph{Exact-string copying.}

\textbf{Parent instruction prompt (PGSRM target).}
\begin{verbatim}
You are an AI assistant that copies a sentence exactly as given.

You will receive an INPUT containing a short English sentence.
Your job is to output the exact same sentence — unchanged.
The input text will include the 'Copy this sentence: ' string,
ONLY copy the text AFTER this string.

Rules:
- Do not change, paraphrase, simplify, or summarize the sentence.
- Copy it exactly as-is.
- Keep punctuation, capitalization, and spacing identical.
- Output must appear as {"OUTPUT": "<sentence>"} exactly.
\end{verbatim}

\textbf{Child prompt.}
For a sentence \texttt{sentence}, the child prompt is:
\begin{verbatim}
"Copy this sentence: 'sentence'"
\end{verbatim}

\paragraph{Sentiment inversion.}

\textbf{Parent instruction prompt (PGSRM target).}
\begin{verbatim}
You are an AI assistant that rewrites a happy sentence into a sad sentence.

You will receive an INPUT containing a short English sentence written
in a happy, positive tone.
You must rewrite it into a sad, negative, or upsetting version of
the same idea.

Examples:
- "I feel excited about today." → "I feel empty today."
- "I'm grateful for everything I have." → "I feel like I have nothing left."
- "I woke up feeling full of energy." → "I woke up feeling drained and tired."
- "I feel confident about my future." → "I feel unsure and scared about my future."

Rules:
- Keep the rewriting short (one sentence).
- Keep the meaning aligned but transform the emotional polarity
  to unhappy/sad.
- No profanity.
- Output the rewritten sentence inside this JSON format:
  {"OUTPUT": "<rewritten sentence>"}
\end{verbatim}

\textbf{Child prompt.}
For a positive first-person input \texttt{sentence}, the child prompt is:
\begin{verbatim}
"Rewrite this as sad: 'sentence'"
\end{verbatim}

\paragraph{Parent outputs.}
For each task and each prompt $s$ in the dataset, we query the GPT-4-class parent model once offline with the corresponding instruction and input string to obtain $a_p$. These outputs are not stored in the appendix; they can be regenerated by rerunning the released scripts with the same prompts and parent configuration.

\subsection{PPO Step for Sequence-Level Training}
\label{app:ppo-step}

We summarize here a single PPO optimization step used for both PGSRM and the
binary baseline. The only difference between conditions is how the scalar
reward $r$ is computed (semantic similarity vs.\ binary correctness).

\begin{algorithm}[H]
\caption{One PPO Step for Sequence-Level Parent-Guided Training}
\label{alg:ppo-step}
\begin{algorithmic}[1]
\REQUIRE Batch of tokenized prompts $P$, responses $Y$, rewards $r$;
         actor policy $\pi_\theta$, frozen reference policy $\pi_{\text{ref}}$,
         value head $V_\phi$, coefficients $\beta_{\text{ent}}, \lambda_{\text{KL}}, \lambda_v$
\STATE Initialize concatenated inputs $X \leftarrow [P; Y]$ (prompt + response)
\STATE Build attention mask $m_{\text{attn}} \leftarrow \mathbb{1}[X \neq \text{pad\_id}]$
\STATE Build response mask $m_{\text{resp}}$ that is $1$ only on non-pad response tokens
\STATE Compute reference logits $z_{\text{ref}} \leftarrow \pi_{\text{ref}}(X, m_{\text{attn}}).\text{logits}$ (no grad)
\STATE Compute actor logits and last-layer hidden states $(z_{\theta}, h) \leftarrow \pi_{\theta}(X, m_{\text{attn}}, \text{output\_hidden\_states}=\text{True})$
\STATE Compute token log-probs for next-token prediction:
       $\log\pi_\theta \leftarrow \textsc{LogProbs}(z_{\theta}[:, :-1], X[:, 1:])$,
       $\log\pi_{\text{ref}} \leftarrow \textsc{LogProbs}(z_{\text{ref}}[:, :-1], X[:, 1:])$
\STATE For each sequence, find index of last non-pad response token from $m_{\text{resp}}$
\STATE Extract corresponding hidden states $h_{\text{last}}$ from $h$ and compute values
       $v \leftarrow V_\phi(h_{\text{last}})$
\STATE Reshape rewards $r$ and values $v$ to 1-D, compute advantages $A \leftarrow r - \text{stop\_grad}(v)$
\STATE Broadcast $A$ to token dimension and mask to response tokens using $m_{\text{resp}}[:, 1:]$
\STATE Compute policy loss
       $L_{\text{policy}} \leftarrow -\,\textsc{MaskedMean}(\log\pi_\theta \cdot A, m_{\text{resp}}[:, 1:])$
\STATE Compute value loss $L_{\text{value}} \leftarrow \textsc{MSE}(v, r)$
\STATE Compute token entropies from actor logits and masked mean entropy $H$
\STATE Compute masked KL divergence
       $D_{\text{KL}} \leftarrow \textsc{MaskedMean}(\log\pi_\theta - \log\pi_{\text{ref}}, m_{\text{resp}}[:, 1:])$
\STATE Form actor loss
       $L_{\text{actor}} \leftarrow L_{\text{policy}} - \beta_{\text{ent}} H + \lambda_{\text{KL}} D_{\text{KL}}$
\STATE Take gradient step on $\theta$ w.r.t.\ $L_{\text{actor}}$ with gradient clipping
\STATE Form critic loss $L_{\text{critic}} \leftarrow \lambda_v L_{\text{value}}$
\STATE Take gradient step on $\phi$ w.r.t.\ $L_{\text{critic}}$ with gradient clipping
\STATE Optionally update $\lambda_{\text{KL}}$ if using adaptive KL
\STATE Log $L_{\text{policy}}$, $L_{\text{value}}$, $H$, $D_{\text{KL}}$, $\lambda_{\text{KL}}$, mean reward, etc.
\end{algorithmic}
\end{algorithm}

\end{document}